\documentclass[10pt,twocolumn,letterpaper]{article}

\usepackage{cvpr}
\usepackage{times}
\usepackage{epsfig}
\usepackage{subfigure}
\usepackage{graphicx}
\usepackage{amssymb}
\usepackage{mathrsfs} 
\usepackage{amsmath,bm}
\usepackage{array,multirow}
\usepackage{mdwlist}
\usepackage{paralist}
\usepackage{url}
\usepackage{color}
\usepackage{caption}
\usepackage{setspace}
\usepackage{textcomp}
\usepackage{bbding}
\usepackage{booktabs}
\usepackage{threeparttable}
\usepackage{diagbox}
\usepackage{enumitem}
\usepackage{xcolor}
\usepackage{algorithm}
\usepackage{algorithmic}

\def\eg{{\em e.g.}}
\def\ie{{\em i.e.}}
\def\etal{{\em et al.}}

\usepackage[pagebackref=true,breaklinks=true,letterpaper=true,colorlinks,bookmarks=false]{hyperref}

\cvprfinalcopy 

\def\cvprPaperID{106}

\ifcvprfinal\fi
\begin{document}

\title{Bridging the Gap Between Anchor-based and Anchor-free Detection via Adaptive Training Sample Selection}

\author{Shifeng Zhang$^{1,2}$, Cheng Chi$^{3}$, Yongqiang Yao$^4$, Zhen Lei$^{1,2}$\thanks{Corresponding author},\ \ Stan Z. Li$^{5}$\\
$^1$ CBSR, NLPR, CASIA\quad
$^2$ SAI, UCAS\quad
$^3$ AIR, CAS\quad
$^4$ BUPT\quad
$^5$ Westlake University \\
{\tt\small \{shifeng.zhang,zlei,szli\}@nlpr.ia.ac.cn, chicheng15@mails.ucas.ac.cn, yao\_yq@bupt.edu.cn}}

\maketitle
\thispagestyle{empty}

\begin{abstract}
Object detection has been dominated by anchor-based detectors for several years. Recently, anchor-free detectors have become popular due to the proposal of FPN and Focal Loss. In this paper, we first point out that the essential difference between anchor-based and anchor-free detection is actually \textbf{how to define positive and negative training samples}, which leads to the performance gap between them. If they adopt the same definition of positive and negative samples during training, there is no obvious difference in the final performance, no matter regressing from a box or a point. This shows that how to select positive and negative training samples is important for current object detectors. Then, we propose an Adaptive Training Sample Selection (ATSS) to automatically select positive and negative samples according to statistical characteristics of object. It significantly improves the performance of anchor-based and anchor-free detectors and bridges the gap between them. Finally, we discuss the necessity of tiling multiple anchors per location on the image to detect objects. Extensive experiments conducted on MS COCO support our aforementioned analysis and conclusions. With the newly introduced ATSS, we improve state-of-the-art detectors by a large margin to $50.7\%$ AP without introducing any overhead. The code is available at \url{https://github.com/sfzhang15/ATSS}.
\end{abstract}

\section{Introduction}
Object detection is a long-standing topic in the field of computer vision, aiming to detect objects of predefined categories. Accurate object detection would have far reaching impact on various applications including image recognition and video surveillance. In recent years, with the development of convolutional neural network (CNN), object detection has been dominated by anchor-based detectors, which can be generally divided into one-stage methods \cite{DBLP:conf/eccv/LiuAESRFB16,DBLP:conf/iccv/LinPRK17} and two-stage methods \cite{DBLP:journals/pami/RenHG017,DBLP:conf/nips/DaiLHS16}. Both of them first tile a large number of preset anchors on the image, then predict the category and refine the coordinates of these anchors by one or several times, finally output these refined anchors as detection results. Because two-stage methods refine anchors several times more than one-stage methods, the former one has more accurate results while the latter one has higher computational efficiency. State-of-the-art results on common detection benchmarks are still held by anchor-based detectors.

Recent academic attention has been geared toward anchor-free detectors due to the emergence of FPN \cite{DBLP:conf/cvpr/LinDGHHB17} and Focal Loss \cite{DBLP:conf/iccv/LinPRK17}. Anchor-free detectors directly find objects without preset anchors in two different ways. One way is to first locate several pre-defined or self-learned keypoints and then bound the spatial extent of objects. We call this type of anchor-free detectors as keypoint-based methods \cite{DBLP:conf/eccv/LawD18, DBLP:conf/cvpr/ZhouZK19}. Another way is to use the center point or region of objects to define positives and then predict the four distances from positives to the object boundary. We call this kind of anchor-free detectors as center-based methods \cite{DBLP:conf/iccv/abs-1904-01355, DBLP:journals/corr/abs-1904-03797}. These anchor-free detectors are able to eliminate those hyperparameters related to anchors and have achieved similar performance with anchor-based detectors, making them more potential in terms of generalization ability.

Among these two types of anchor-free detectors, keypoint-based methods follow the standard keypoint estimation pipeline that is different from anchor-based detectors. However, center-based detectors are similar to anchor-based detectors, which treat points as preset samples instead of anchor boxes. Take the one-stage anchor-based detector RetinaNet \cite{DBLP:conf/iccv/LinPRK17} and the center-based anchor-free detector FCOS \cite{DBLP:conf/iccv/abs-1904-01355} as an example, there are three main differences between them: (1) The number of anchors tiled per location. RetinaNet tiles several anchor boxes per location, while FCOS tiles one anchor point\footnote{A point in FCOS is equal to the center of an anchor box in RetinaNet, thus we call it as the anchor point. A pair of anchor point and box is associated to the same location of feature map to be classified and regressed.} per location. (2) The definition of positive and negative samples. RetinaNet resorts to the Intersection over Union (IoU) for positives and negatives, while FCOS utilizes spatial and scale constraints to select samples. (3) The regression starting status. RetinaNet regresses the object bounding box from the preset anchor box, while FCOS locates the object from the anchor point. As reported in \cite{DBLP:conf/iccv/abs-1904-01355}, the anchor-free FCOS achieves much better performance than the anchor-based RetinaNet, it is worth studying which of these three differences are essential factors for the performance gap.

In this paper, we investigate the differences between anchor-based and anchor-free methods in a fair way by strictly ruling out all the implementation inconsistencies between them. It can be concluded from experiment results that the essential difference between these two kind of methods is the definition of positive and negative training samples, which results in the performance gap between them. If they select the same positive and negative samples during training, there is no obvious gap in the final performance, no matter regressing from a box or a point. Therefore, how to select positive and negative training samples deserves further study. Inspired by that, we propose a new Adaptive Training Sample Selection (ATSS) to automatically select positive and negative samples based on object characteristics. It bridges the gap between anchor-based and anchor-free detectors. Besides, through a series of experiments, a conclusion can be drawn that tiling multiple anchors per location on the image to detect objects is not necessary. Extensive experiments on the MS COCO \cite{DBLP:conf/eccv/LinMBHPRDZ14} dataset support our analysis and conclusions. State-of-the-art AP $50.7\%$ is achieved by applying the newly introduced ATSS without introducing any overhead. The main contributions of this work can be summarized as:

\begin{itemize}
\setlength{\itemsep}{0pt}
\item Indicating the essential difference between anchor-based and anchor-free detectors is actually how to define positive and negative training samples.
\item Proposing an adaptive training sample selection to automatically select positive and negative training samples according to statistical characteristics of object. 
\item Demonstrating that tiling multiple anchors per location on the image to detect objects is a useless operation.
\item Achieving state-of-the-art performance on MS COCO without introducing any additional overhead.
\end{itemize}

\section{Related Work}
Current CNN-based object detection consists of anchor-based and anchor-free detectors. The former one can be divided into two-stage and one-stage methods, while the latter one falls into keypoint-based and center-based methods.

\subsection{Anchor-based Detector}
{\noindent \textbf{Two-stage method.}} The emergence of Faster R-CNN \cite{DBLP:journals/pami/RenHG017} establishes the dominant position of two-stage anchor-based detectors. Faster R-CNN consists of a separate region proposal network (RPN) and a region-wise prediction network (R-CNN) \cite{DBLP:conf/cvpr/GirshickDDM14,DBLP:conf/iccv/Girshick15} to detect objects. After that, lots of algorithms are proposed to improve its performance, including architecture redesign and reform \cite{DBLP:conf/eccv/CaiFFV16,DBLP:conf/nips/DaiLHS16,DBLP:conf/cvpr/abs-1712-00726,DBLP:journals/corr/LeeEK17,DBLP:journals/corr/abs-1901-01892}, context and attention mechanism \cite{DBLP:conf/cvpr/BellZBG16,DBLP:conf/eccv/ShrivastavaG16,DBLP:journals/corr/abs-1807-00119,DBLP:conf/eccv/ChenHT18,DBLP:journals/corr/abs-1903-11752}, multi-scale training and testing \cite{DBLP:journals/corr/abs-1711-08189,DBLP:journals/corr/abs-1812-01600}, training strategy and loss function \cite{DBLP:conf/cvpr/NajibiRD16,DBLP:conf/cvpr/ShrivastavaGG16,DBLP:conf/cvpr/WangSG17, DBLP:conf/cvpr/HeZWS019}, feature fusion and enhancement \cite{DBLP:conf/cvpr/KongYCS16, DBLP:conf/cvpr/LinDGHHB17}, better proposal and balance \cite{tan2019learning, DBLP:conf/cvpr/PangCSFOL19}. Nowadays, state-of-the-art results are still held by two-stage anchor-based methods on standard detection benchmarks.

{\noindent \textbf{One-stage method.}} With the advent of SSD \cite{DBLP:conf/eccv/LiuAESRFB16}, one-stage anchor-based detectors have attracted much attention because of their high computational efficiency. SSD spreads out anchor boxes on multi-scale layers within a ConvNet to directly predict object category and anchor box offsets. Thereafter, plenty of works are presented to boost its performance in different aspects, such as fusing context information from different layers \cite{DBLP:conf/cvpr/KongSYLLC17, DBLP:journals/corr/FuLRTB17,zhou2018scale}, training from scratch \cite{DBLP:conf/iccv/abs-1708-01241, DBLP:journals/corr/abs-1810-08425}, introducing new loss function \cite{DBLP:conf/iccv/LinPRK17, DBLP:conf/cvpr/ChenLLSWD0HZ19}, anchor refinement and matching \cite{DBLP:conf/cvpr/abs-1711-06897, zhang2019freeanchor}, architecture redesign \cite{DBLP:conf/eccv/KimKSKK18, DBLP:conf/eccv/KongSHL18}, feature enrichment and alignment \cite{DBLP:conf/eccv/LiuHW18, DBLP:journals/corr/abs-1712-00433, wang2019learning, nie2019enriched, li2019dynamic}. At present, one-stage anchor-based methods can achieve very close performance with two-stage anchor-based methods at a faster inference speed.

\subsection{Anchor-free Detector}
{\noindent \textbf{Keypoint-based method.}} This type of anchor-free method first locates several pre-defined or self-learned keypoints, and then generates bounding boxes to detect objects. CornerNet \cite{DBLP:conf/eccv/LawD18} detects an object bounding box as a pair of keypoints (top-left corner and bottom-right corner) and CornerNet-Lite \cite{DBLP:journals/corr/abs-1904-08900} introduces CornerNet-Saccade and CornerNet-Squeeze to improve its speed. The second stage of Grid R-CNN \cite{DBLP:conf/cvpr/LuLYLY19} locates objects via predicting grid points with the position sensitive merits of FCN and then determining the bounding box guided by the grid. ExtremeNet \cite{DBLP:conf/cvpr/ZhouZK19} detects four extreme points (top-most, leftmost, bottom-most, right-most) and one center point to generate the object bounding box. Zhu \etal~\cite{DBLP:journals/corr/abs-1904-07850} use keypoint estimation to find center point of objects and regress to all other properties including size, 3D location, orientation and pose. CenterNet \cite{DBLP:journals/corr/abs-1904-08189} extends CornetNet as a triplet rather than a pair of keypoints to improve both precision and recall. RepPoints \cite{DBLP:journals/corr/abs-1904-11490} represents objects as a set of sample points and learns to arrange themselves in a manner that bounds the spatial extent of an object and indicates semantically significant local areas.

{\noindent \textbf{Center-based method.}} This kind of anchor-free method regards the center (\eg, the center point or part) of object as foreground to define positives, and then predicts the distances from positives to the four sides of the object bounding box for detection. YOLO \cite{DBLP:conf/cvpr/RedmonDGF16} divides the image into an $S\times S$ grid, and the grid cell that contains the center of an object is responsible for detecting this object. DenseBox \cite{DBLP:journals/corr/HuangYDY15} uses a filled circle located in the center of the object to define positives and then predicts the four distances from positives to the bound of the object bounding box for location. GA-RPN \cite{DBLP:conf/cvpr/WangCYLL19} defines the pixels in the center region of the object as positives to predict the location, width and height of object proposals for Faster R-CNN. FSAF \cite{DBLP:conf/cvpr/ZhuHS19} attaches an anchor-free branch with online feature selection to RetinaNet. The newly added branch defines the center region of the object as positives to locate it via predicting four distances to its bounds. FCOS \cite{DBLP:conf/iccv/abs-1904-01355} regards all the locations inside the object bounding box as positives with four distances and a novel centerness score to detect objects. CSP \cite{DBLP:conf/cvpr/LiuLRHY19} only defines the center point of the object box as positives to detect pedestrians with fixed aspect ratio. FoveaBox \cite{DBLP:journals/corr/abs-1904-03797} regards the locations in the middle part of object as positives with four distances to perform detection.

\section{Difference Analysis of Anchor-based and Anchor-free Detection}

Without loss of generality, the representative anchor-based RetinaNet \cite{DBLP:conf/iccv/LinPRK17} and anchor-free FCOS \cite{DBLP:conf/iccv/abs-1904-01355} are adopted to dissect their differences. In this section, we focus on the last two differences: the positive/negative sample definition and the regression starting status. The remaining one difference: the number of anchors tiled per location, will be discussed in subsequent section. Thus, we just tile one square anchor per location for RetinaNet, which is quite similar to FCOS. In the remaining part, we first introduce the experiment settings, then rule out all the implementation inconsistencies, finally point out the essential difference between anchor-based and anchor-free detectors.

\subsection{Experiment Setting}

{\noindent \textbf{Dataset.}} All experiments are conducted on the challenging MS COCO \cite{DBLP:conf/eccv/LinMBHPRDZ14} dataset that includes $80$ object classes. Following the common practice \cite{DBLP:conf/iccv/LinPRK17, DBLP:conf/iccv/abs-1904-01355}, all $115K$ images in the {\tt trainval35k} split is used for training, and all $5K$ images in the {\tt minival} split is used as validation for analysis study. We also submit our main results to the evaluation server for the final performance on the {\tt test-dev} split.

{\noindent \textbf{Training Detail.}} We use the ImageNet \cite{DBLP:journals/ijcv/RussakovskyDSKS15} pretrained ResNet-50 \cite{DBLP:conf/cvpr/HeZRS16} with 5-level feature pyramid structure as the backbone. The newly added layers are initialized in the same way as in \cite{DBLP:conf/iccv/LinPRK17}. For RetinaNet, each layer in the 5-level feature pyramid is associated with one square anchor with $8S$ scale, where $S$ is the total stride size. During training, we resize the input images to keep their shorter side being $800$ and their longer side less or equal to $1,333$. The whole network is trained using the Stochastic Gradient Descent (SGD) algorithm for $90K$ iterations with $0.9$ momentum, $0.0001$ weight decay and $16$ batch size. We set the initial learning rate as $0.01$ and decay it by $0.1$ at iteration $60K$ and $80K$, respectively. Unless otherwise stated, the aforementioned training details are used in the experiments.

{\noindent \textbf{Inference Detail.}} During the inference phase, we resize the input image in the same way as in the training phase, and then forward it through the whole network to output the predicted bounding boxes with a predicted class. After that, we use the preset score $0.05$ to filter out plenty of background bounding boxes, and then output the top $1000$ detections per feature pyramid. Finally, the Non-Maximum Suppression (NMS) is applied with the IoU threshold $0.6$ per class to generate final top $100$ confident detections per image.

\begin{table}[t]
\centering
\caption{Analysis of implementation inconsistencies between RetinaNet and FCOS on MS COCO {\tt minival} set. ``\#A=1'' means there is one square anchor box per location.}
\setlength{\tabcolsep}{3.1pt}
\begin{tabular}{|c|c|cccccc|}
\hline
Inconsistency & FCOS & \multicolumn{6}{c|}{RetinaNet (\#A=1)} \\
\hline
GroupNorm & \Checkmark & & \Checkmark & \Checkmark & \Checkmark & \Checkmark & \Checkmark \\
GIoU Loss & \Checkmark & & & \Checkmark & \Checkmark & \Checkmark & \Checkmark \\
In GT Box & \Checkmark & & & & \Checkmark & \Checkmark & \Checkmark \\
Centerness & \Checkmark & & & & & \Checkmark & \Checkmark \\
Scalar & \Checkmark & & & & & & \Checkmark \\
\hline
AP ($\%$) & 37.8 & 32.5 & 33.4 & 34.9 & 35.3 & 36.8 & 37.0 \\
\hline
\end{tabular}
\label{tab:inconsistency}
\end{table}

\subsection{Inconsistency Removal}

We mark the anchor-based detector RetinaNet with only one square anchor box per location as \textbf{RetinaNet (\#A=1)}, which is almost the same as the anchor-free detector FCOS. However, as reported in \cite{DBLP:conf/iccv/abs-1904-01355}, FCOS outperforms RetinaNet (\#A=1) by a large margin in AP performance on the MS COCO {\tt minival} subset, \ie, $37.1\%$ \emph{vs.} $32.5\%$. Furthermore, some new improvements have been made for FCOS including moving centerness to regression branch, using GIoU loss function and normalizing regression targets by corresponding strides. These improvements boost the AP performance of FCOS from $37.1\%$ to $37.8\%$ \footnote{This $37.8\%$ AP result does not include the center sample improvement, which is our contribution that has been merged into FCOS and will be introduced in Sec. 4.2.}, making the gap even bigger. However, part of the AP gap between the anchor-based detector ($32.5\%$) and the anchor-free detector ($37.8\%$) results from some universal improvements that are proposed or used in FCOS, such as adding GroupNorm \cite{DBLP:conf/eccv/WuH18} in heads, using the GIoU \cite{DBLP:conf/cvpr/RezatofighiTGS019} regression loss function, limiting positive samples in the ground-truth box \cite{DBLP:conf/iccv/abs-1904-01355}, introducing the centerness branch \cite{DBLP:conf/iccv/abs-1904-01355} and adding a trainable scalar \cite{DBLP:conf/iccv/abs-1904-01355} for each level feature pyramid. These improvements can also be applied to anchor-based detectors, therefore they are not the essential differences between anchor-based and anchor-free methods. We apply them to RetinaNet (\#A=1) one by one so as to rule out these implementation inconsistencies. As listed in Table \ref{tab:inconsistency}, these irrelevant differences improve the anchor-based RetinaNet to $37.0\%$, which still has a gap of $0.8\%$ to the anchor-free FCOS. By now, after removing all the irrelevant differences, we can explore the essential differences between anchor-based and anchor-free detectors in a quite fair way.

\begin{figure}[t]
\centering
\includegraphics[width=1.0\linewidth]{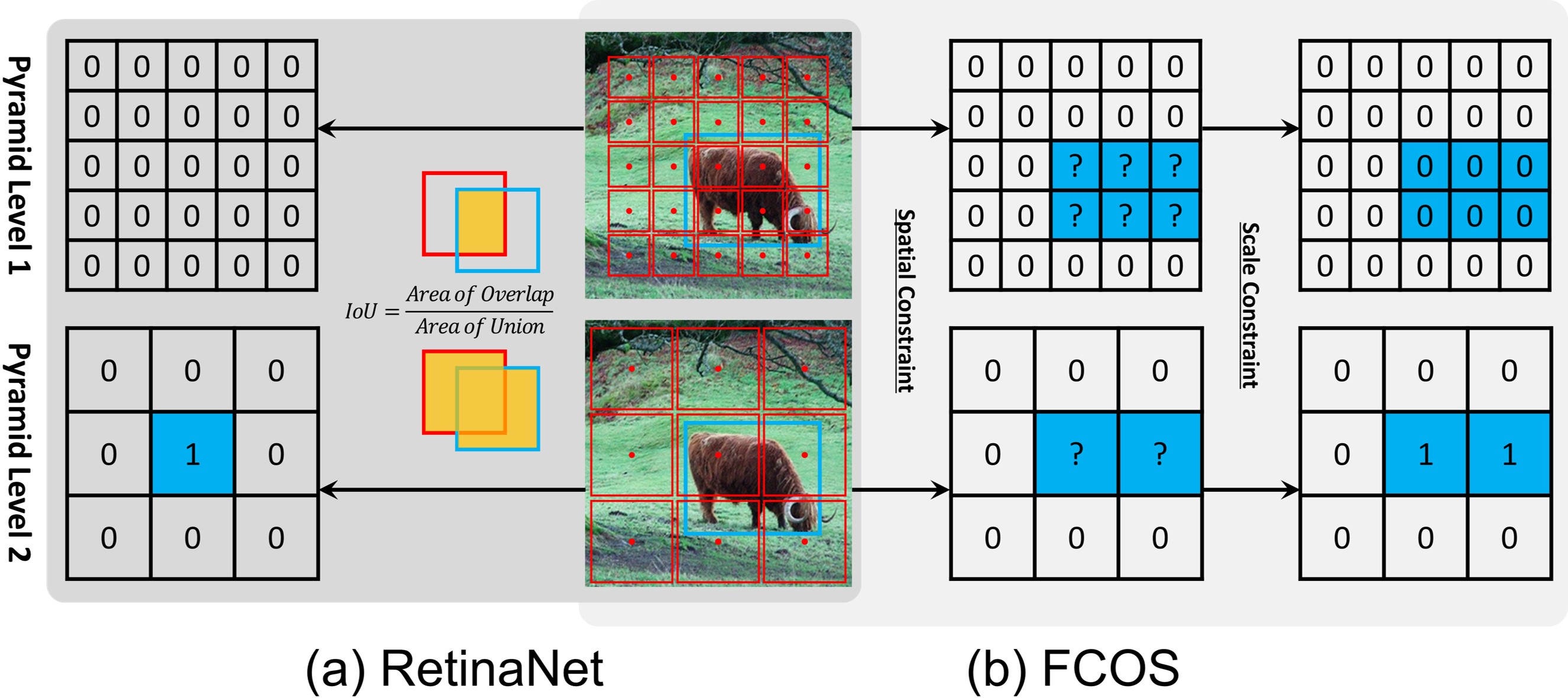}
\caption{Definition of positives (\colorbox{cyan}{\tiny{1}}) and negatives (\colorbox{lightgray}{\tiny{0}}). Blue box, red box and red point are ground-truth, anchor box and anchor point. (a) RetinaNet uses IoU to select positives (\colorbox{cyan}{\tiny{1}}) in spatial and scale dimension simultaneously. (b) FCOS first finds candidate positives (\colorbox{cyan}{\tiny{?}}) in spatial dimension, then selects final positives (\colorbox{cyan}{\tiny{1}}) in scale dimension.}
\label{fig:cls}
\end{figure}

\subsection{Essential Difference}

After applying those universal improvements, these are only two differences between the anchor-based RetinaNet (\#A=1) and the anchor-free FCOS. One is about the classification sub-task in detection, \ie, the way to define positive and negative samples. Another one is about the regression sub-task, \ie, the regression starting from an anchor box or an anchor point.

{\noindent \textbf{Classification.}} As shown in Figure \ref{fig:cls}(a), RetinaNet utilizes IoU to divide the anchor boxes from different pyramid levels into positives and negatives. It first labels the best anchor box of each object and the anchor boxes with IoU $> \theta_{p}$ as positives, then regards the anchor boxes with IoU $< \theta_{n}$ as negatives, finally other anchor boxes are ignored during training. As shown in Figure \ref{fig:cls}(b), FCOS uses spatial and scale constraints to divide the anchor points from different pyramid levels. It first considers the anchor points within the ground-truth box as candidate positive samples, then selects the final positive samples from candidates based on the scale range defined for each pyramid level\footnote{There are several preset hyperparameters in FCOS to define the scale range for five pyramid levels: [$m2$, $m3$] for P3, [$m3$, $m4$] for P4, [$m4$, $m5$] for P5, [$m5$, $m6$] for P6 and [$m6$, $m7$] for P7.}, finally those unselected anchor points are negative samples.

As shown in Figure \ref{fig:cls}, FCOS first uses the spatial constraint to find candidate positives in the spatial dimension, then uses the scale constraint to select final positives in the scale dimension. In contrast, RetinaNet utilizes IoU to directly select the final positives in the spatial and scale dimension simultaneously. These two different sample selection strategies produce different positive and negative samples. As listed in the first column of Table \ref{tab:difference} for RetinaNet (\#A=1), using the spatial and scale constraint strategy instead of the IoU strategy improves the AP performance from $37.0\%$ to $37.8\%$. As for FCOS, if it uses the IoU strategy to select positive samples, the AP performance decreases from $37.8\%$ to $36.9\%$ as listed in the second column of Table \ref{tab:difference}. These results demonstrate that the definition of positive and negative samples is an essential difference between anchor-based and anchor-free detectors.

\begin{figure}[t]
\centering
\subfigure[Positive sample]{
\label{fig:reg1}
\includegraphics[width=0.3\linewidth]{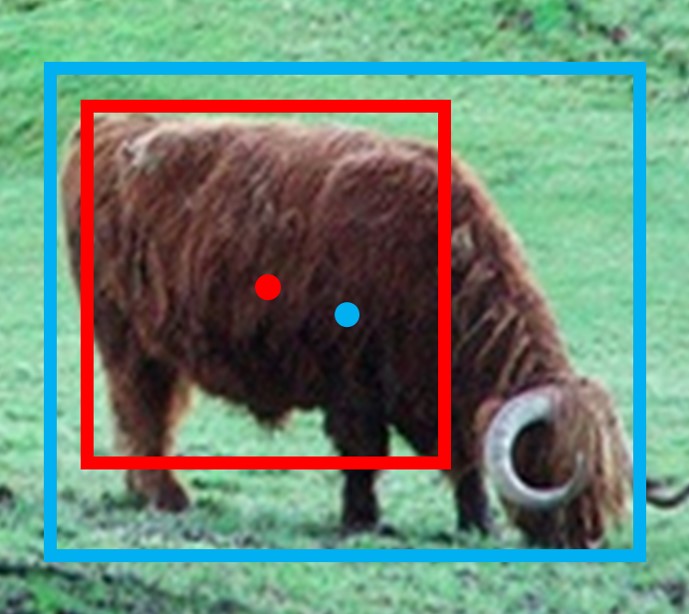}}
\subfigure[RetinaNet]{
\label{fig:reg2}
\includegraphics[width=0.3\linewidth]{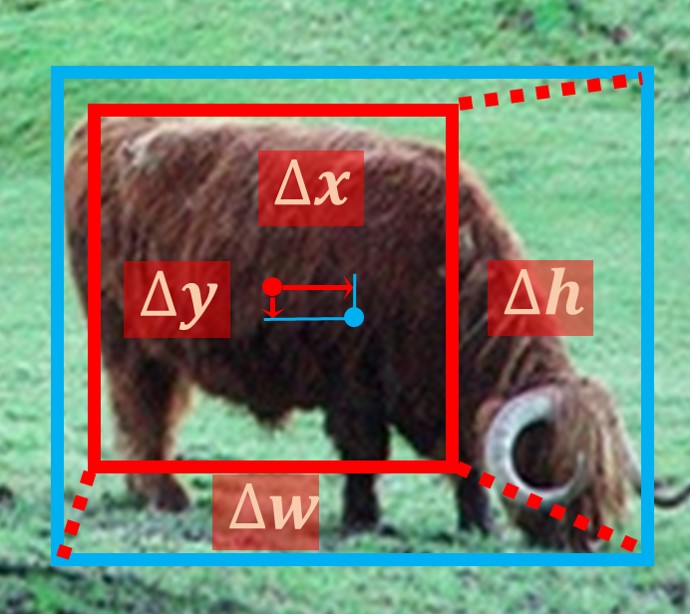}}
\subfigure[FCOS]{
\label{fig:reg3}
\includegraphics[width=0.3\linewidth]{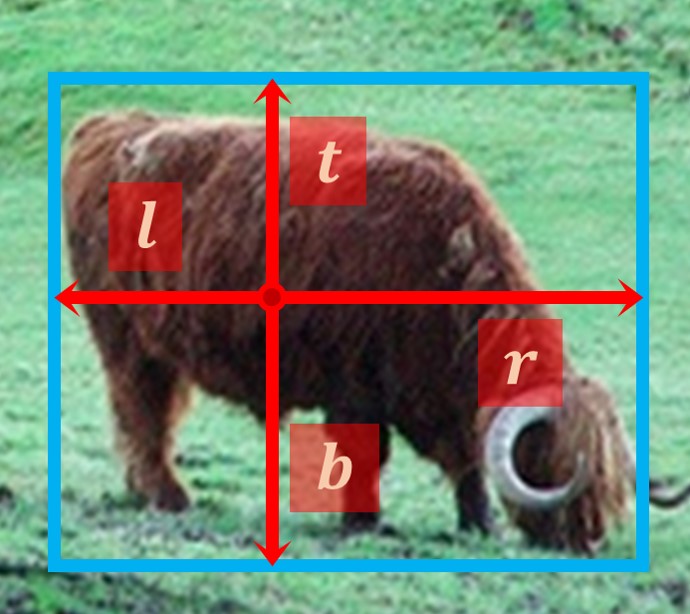}}
\caption{(a) Blue point and box are the center and bound of object, red point and box are the center and bound of anchor. (b) RetinaNet regresses from anchor box with four offsets. (c) FCOS regresses from anchor point with four distances.}
\label{fig:regression}
\end{figure} 

\begin{table}[t]
\renewcommand\arraystretch{0.96}
\centering
\caption{Analysis of differences ($\%$) between RetinaNet and FCOS on the MS COCO {\tt minival} set.}
\setlength{\tabcolsep}{11.0pt}
\begin{tabular}{|l|c|c|}
\hline
\diagbox{Classification}{Regression} & Box & Point \\
\hline
Intersection over Union       &37.0 &36.9 \\
Spatial and Scale Constraint  &37.8 &37.8 \\
\hline
\end{tabular}
\label{tab:difference}
\end{table}

{\noindent \textbf{Regression.}} After positive and negative samples are determined, the location of object is regressed from positive samples as shown in Figure \ref{fig:regression}(a). RetinaNet regresses from the anchor box with four offsets between the anchor box and the object box as shown in Figure \ref{fig:regression}(b), while FCOS regresses from the anchor point with four distances to the bound of object as shown in Figure \ref{fig:regression}(c). It means that for a positive sample, the regression starting status of RetinaNet is a box while FCOS is a point. However, as shown in the first and second rows of Table \ref{tab:difference}, when RetinaNet and FCOS adopt the same sample selection strategy to have consistent positive/negative samples, there is no obvious difference in final performance, no matter regressing starting from a point or a box, \ie, $37.0\%$ \emph{vs.} $36.9\%$ and $37.8\%$ \emph{vs.} $37.8\%$. These results indicate that the regression starting status is an irrelevant difference rather than an essential difference.

{\noindent \textbf{Conclusion.}} According to these experiments conducted in a fair way, we indicate that the essential difference between one-stage anchor-based detectors and center-based anchor-free detectors is actually how to define positive and negative training samples, which is important for current object detection and deserves further study.

\section{Adaptive Training Sample Selection}

When training an object detector, we first need to define positive and negative samples for classification, and then use positive samples for regression. According to the previous analysis, the former one is crucial and the anchor-free detector FCOS improves this step. It introduces a new way to define positives and negatives, which achieves better performance than the traditional IoU-based strategy. Inspired by this, we delve into the most basic issue in object detection: \emph{how to define positive and negative training samples}, and propose an Adaptive Training Sample Selection (ATSS). Compared with these traditional strategies, our method almost has no hyperparameters and is robust to different settings.

\subsection{Description}

Previous sample selection strategies have some sensitive hyperparameters, such as IoU thresholds in anchor-based detectors and scale ranges in anchor-free detectors. After these hyperparameters are set, all ground-truth boxes must select their positive samples based on the fixed rules, which are suitable for most objects, but some outer objects will be neglected. Thus, different settings of these hyperparameters will have very different results.

To this end, we propose the ATSS method that automatically divides positive and negative samples according to statistical characteristics of object almost without any hyperparameter. Algorithm \ref{alg:atss} describes how the proposed method works for an input image. For each ground-truth box $g$ on the image, we first find out its candidate positive samples. As described in Line $3$ to $6$, on each pyramid level, we select $k$ anchor boxes whose center are closest to the center of $g$ based on L2 distance. Supposing there are $\mathcal{L}$ feature pyramid levels, the ground-truth box $g$ will have $k\times\mathcal{L}$ candidate positive samples. After that, we compute the IoU between these candidates and the ground-truth $g$ as $\mathcal{D}_g$ in Line $7$, whose mean and standard deviation are computed as $m_g$ and $v_g$ in Line $8$ and Line $9$. With these statistics, the IoU threshold for this ground-truth $g$ is obtained as $t_g=m_g+v_g$ in Line $10$. Finally, we select these candidates whose IoU are greater than or equal to the threshold $t_g$ as final positive samples in Line $11$ to $15$. Notably, we also limit the positive samples' center to the ground-truth box as shown in Line $12$. Besides, if an anchor box is assigned to multiple ground-truth boxes, the one with the highest IoU will be selected. The rest are negative samples. Some motivations behind our method are explained as follows.

\begin{algorithm}[t!]
\small
\caption{Adaptive Training Sample Selection (ATSS)} 
\label{alg:atss} 
\begin{algorithmic}[1]
\REQUIRE ~~\\
$\mathcal{G}$ is a set of ground-truth boxes on the image \\
$\mathcal{L}$ is the number of feature pyramid levels \\
$\mathcal{A}_{i}$ is a set of anchor boxes from the $i_{th}$ pyramid levels \\
$\mathcal{A}$ is a set of all anchor boxes \\
$k$ is a quite robust hyperparameter with a default value of $9$\\
\ENSURE ~~\\
$\mathcal{P}$ is a set of positive samples \\
$\mathcal{N}$ is a set of negative samples \\
\vspace{2mm}
\FOR{each ground-truth $g \in \mathcal{G}$}
\STATE build an empty set for candidate positive samples of the ground-truth $g$: {$\mathcal{C}_g \leftarrow{} \varnothing $}; \\
\FOR{each level $i \in [1,\mathcal{L}]$}
\STATE $\mathcal{S}_{i}\leftarrow{}$ select $k$ anchors from $A_{i}$ whose center are closest to the center of ground-truth $g$ based on L2 distance; \\
\STATE $\mathcal{C}_g=\mathcal{C}_g \cup \mathcal{S}_{i};$ \\
\ENDFOR
\STATE compute IoU between $\mathcal{C}_g$ and $g$: $\mathcal{D}_g=IoU(\mathcal{C}_g, g);$ \\
\STATE compute mean of $\mathcal{D}_g$: $m_g=Mean(\mathcal{D}_g)$; \\
\STATE compute standard deviation of $\mathcal{D}_g$: $v_g=Std(\mathcal{D}_g)$; \\
\STATE compute IoU threshold for ground-truth $g$: $t_g=m_g+v_g$; \\
\FOR{each candidate $c \in \mathcal{C}_g$}
\IF{ $IoU(c, g) \geq t_g$ and center of $c$ in $g$}
\STATE $\mathcal{P}=\mathcal{P} \cup c;$ \\
\ENDIF
\ENDFOR
\ENDFOR
\STATE $\mathcal{N}=\mathcal{A}-\mathcal{P};$ \\
\RETURN $\mathcal{P}, \mathcal{N}$;
\end{algorithmic}
\end{algorithm}

{\noindent \textbf{Selecting candidates based on the center distance between anchor box and object.} For RetinaNet, the IoU is larger when the center of anchor box is closer to the center of object. For FCOS, the closer anchor point to the center of object will produce higher-quality detections. Thus, the closer anchor to the center of object is the better candidate.

{\noindent \textbf{Using the sum of mean and standard deviation as the IoU threshold.} The IoU mean $m_g$ of an object is a measure of the suitability of the preset anchors for this object. A high $m_g$ as shown in Figure \ref{fig:atss}(a) indicates it has high-quality candidates and the IoU threshold is supposed to be high. A low $m_g$ as shown in Figure \ref{fig:atss}(b) indicates that most of its candidates are low-quality and the IoU threshold should be low. Besides, the IoU standard deviation $v_g$ of an object is a measure of which layers are suitable to detect this object. A high $v_g$ as shown in Figure \ref{fig:atss}(a) means there is a pyramid level specifically suitable for this object, adding $v_g$ to $m_g$ obtains a high threshold to select positives only from that level. A low $v_g$ as shown in Figure \ref{fig:atss}(b) means that there are several pyramid levels suitable for this object, adding $v_g$ to $m_g$ obtains a low threshold to select appropriate positives from these levels. Using the sum of mean $m_g$ and standard deviation $v_g$ as the IoU threshold $t_g$ can adaptively select enough positives for each object from appropriate pyramid levels in accordance of statistical characteristics of object.

{\noindent \textbf{Limiting the positive samples' center to object.} The anchor with a center outside object is a poor candidate and will be predicted by the features outside the object, which is not conducive to training and should be excluded.

{\noindent \textbf{Maintaining fairness between different objects.} According to the statistical theory\footnote{\url{http://dwz1.cc/sNIgLI2}}, about $16\%$ of samples are in the confidence interval $[m_g+v_g, 1]$ in theory. Although the IoU of candidates is not a standard normal distribution, the statistical results show that each object has about $0.2*k\mathcal{L}$ positive samples, which is invariant to its scale, aspect ratio and location. In contrast, strategies of RetinaNet and FCOS tend to have much more positive samples for larger objects, leading to unfairness between different objects.

{\noindent \textbf{Keeping almost hyperparameter-free.} Our method only has one hyperparameter $k$. Subsequent experiments prove that it is quite insensitive to the variations of $k$ and the proposed ATSS can be considered almost hyperparameter-free.

\begin{figure}[t]
\centering
\subfigure[]{
\label{fig:atss1}
\includegraphics[width=0.48\linewidth]{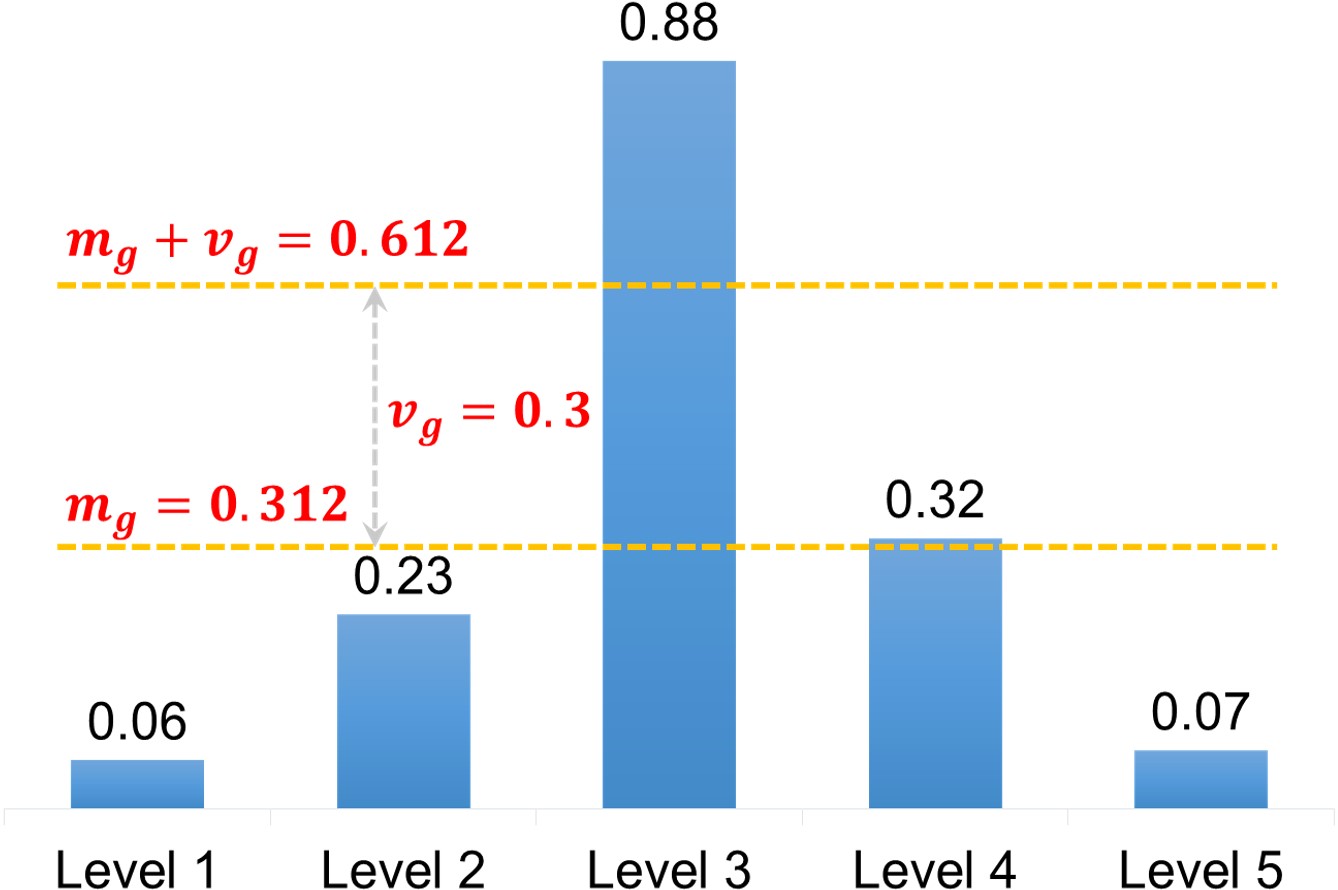}}
\subfigure[]{
\label{fig:atss2}
\includegraphics[width=0.48\linewidth]{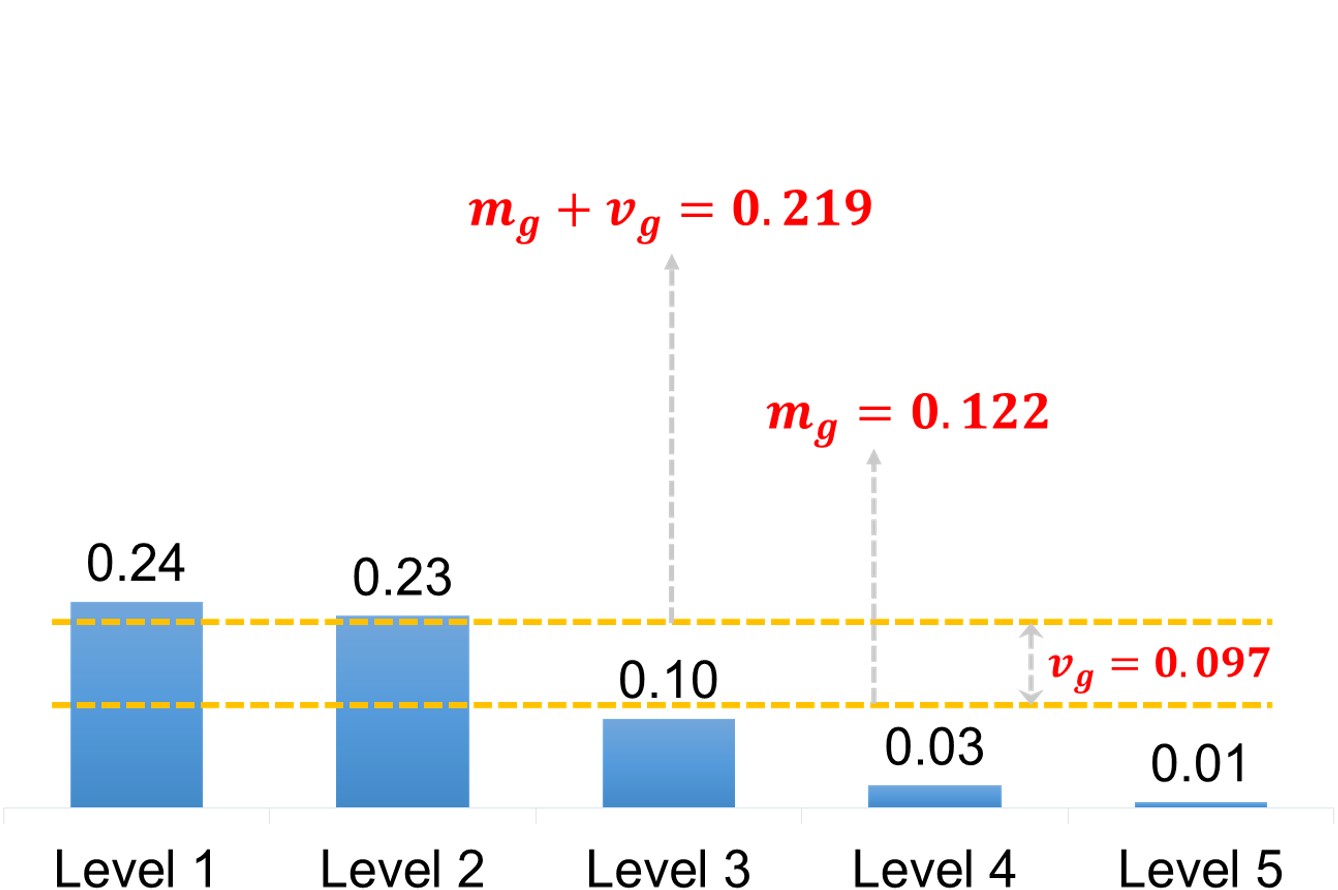}}
\caption{Illustration of ATSS. Each level has one candidate with its IoU. (a) A ground-truth with a high $m_g$ and a high $v_g$. (b) A ground-truth with a low $m_g$ and a low $v_g$.}
\label{fig:atss}
\end{figure} 

\subsection{Verification}

{\noindent \textbf{Anchor-based RetinaNet.}} To verify the effectiveness of our adaptive training sample selection for anchor-based detectors, we use it to replace the traditional strategy in the improved RetinaNet (\#A=1). As shown in Table \ref{tab:verification}, it consistently boosts the performance by $2.3\%$ on AP, $2.4\%$ on AP$_{50}$, $2.9\%$ for AP$_{75}$, $2.9\%$ for AP$_S$, $2.1\%$ for AP$_M$ and $2.7\%$ for AP$_L$. These improvements are mainly due to the adaptive selection of positive samples for each ground-truth based on its statistical characteristics. Since our method only redefines positive and negative samples without incurring any additional overhead, these improvements can be considered cost-free.

{\noindent \textbf{Anchor-free FCOS.}} The proposed method can also be applied to the anchor-free FCOS in two different versions: the lite and full version. For the lite version, we apply some ideas of the proposed ATSS to FCOS, \ie, replacing its way to select candidate positives with the way in our method. FCOS considers anchor points in the object box as candidates, which results in plenty of low-quality positives. In contrast, our method selects top $k=9$ candidates per pyramid level for each ground-truth. The lite version of our method has been merged to the official code of FCOS as the center sampling, which improves FCOS from $37.8\%$ to $38.6\%$ on AP as listed in Table \ref{tab:verification}. However, the hyperparameters of scale ranges still exist in the lite version.

For the full version, we let the anchor point in FCOS become the anchor box with $8S$ scale to define positive and negative samples, then still regress these positive samples to objects from the anchor point like FCOS. As shown in Table \ref{tab:verification}, it significantly increases the performance by $1.4\%$ for AP, by $1.7\%$ for AP$_{50}$, by $1.7\%$ for AP$_{75}$, by $0.6\%$ for AP$_{S}$, by $1.3\%$ for AP$_{M}$ and by $2.7\%$ for AP$_{L}$. Notably, these two versions have the same candidates selected in the spatial dimension, but different ways to select final positives from candidates along the scale dimension. As listed in the last two rows of Table \ref{tab:verification}, the full version (ATSS) outperforms the lite version (center sampling) across different metrics by a large margin. These results indicate that the adaptive way in our method is better than the fixed way in FCOS to select positives from candidates along the scale dimension.

\begin{table}[t]
\centering
\caption{Verification of the proposed method ($\%$) on the MS COCO {\tt minival} set. ATSS and center sampling are the full version and the lite version of our proposed method.}
\footnotesize \setlength{\tabcolsep}{3.5pt}
\begin{tabular}{|l|ccc|ccc|}
\hline
Method & AP & AP$_{50}$ & AP$_{75}$ & AP$_{\it S}$ & AP$_{\it M}$ & AP$_{\it L}$ \\
\hline
RetinaNet (\#A=1)        & 37.0 & 55.1 & 39.9 & 21.4 & 41.2 & 48.6 \\
RetinaNet (\#A=1) + ATSS & 39.3 & 57.5 & 42.8 & 24.3 & 43.3 & 51.3 \\
\hline
FCOS                     & 37.8 & 55.6 & 40.7 & 22.1 & 41.8 & 48.8 \\
FCOS + Center sampling   & 38.6 & 57.4 & 41.4 & 22.3 & 42.5 & 49.8 \\
FCOS + ATSS              & 39.2 & 57.3 & 42.4 & 22.7 & 43.1 & 51.5 \\
\hline
\end{tabular}
\label{tab:verification}
\end{table}

\begin{table}[b]
\centering
\caption{Analysis of different values of hyperparameter $k$ on the MS COCO {\tt minival} set.}
\footnotesize \setlength{\tabcolsep}{4.2pt}
\begin{tabular}{|c|ccccccccc|}
\hline
$k$ & 3 & 5 & 7 & 9 & 11 & 13 & 15 & 17 & 19\\
\hline
AP ($\%$) & 38.0 & 38.8 & 39.1 & 39.3 & 39.1 & 39.0 & 39.1 & 39.2 & 38.9 \\
\hline
\end{tabular}
\label{tab:topk}
\end{table}

\subsection{Analysis}
Training an object detector with the proposed adaptive training sample selection only involves one hyperparameter $k$ and one related setting of anchor boxes. This subsection analyzes them one after another.

{\noindent \textbf{Hyperparameter $k$.}} We conduct several experiments to study the robustness of the hyperparameter $k$, which is used to select the candidate positive samples from each pyramid level. As shown in Table \ref{tab:topk}, different values of $k$ in $[3,5,7,9,11,13,15,17,19]$ are used to train the detector. We observe that the proposed method is quite insensitive to the variations of $k$ from $7$ to $17$. Too large $k$ (\eg, $19$) will result in too many low-quality candidates that slightly decreases the performance. Too small $k$ (\eg, $3$) causes a noticeable drop in accuracy, because too few candidate positive samples will cause statistical instability. Overall, the only hyperparameter $k$ is quite robust and the proposed ATSS can be nearly regarded as hyperparameter-free.

{\noindent \textbf{Anchor Size.}} The introduced method resorts to the anchor boxes to define positives and we also study the effect of the anchor size. In the previous experiments, one square anchor with $8S$ ($S$ indicates the total stride size of the pyramid level) is tiled per location. As shown in Table \ref{tab:anchor_scale}, we conduct some experiments with different scales of the square anchor in $[5,6,7,8,9]$ and the performances are quite stable. Besides, several experiments with different aspect ratios of the $8S$ anchor box are performed as shown in Table \ref{tab:anchor_ratio}. The performances are also insensitive to this variation. These results indicate that the proposed method is robust to different anchor settings.

\begin{table}[t]
\centering
\caption{Analysis ($\%$) of different anchor scales with fixed aspect ratio $1:1$ on the MS COCO {\tt minival} set.}
\footnotesize \setlength{\tabcolsep}{8.5pt}
\begin{tabular}{|c|ccc|ccc|}
\hline
Scale & AP & AP$_{50}$ & AP$_{75}$ & AP$_{\it S}$ & AP$_{\it M}$ & AP$_{\it L}$ \\
\hline
5  & 39.0 & 57.9 & 41.9 & 23.2 & 42.8 & 50.5 \\
6  & 39.2 & 57.6 & 42.5 & 23.5 & 42.8 & 51.1 \\
7  & 39.3 & 57.6 & 42.4 & 22.9 & 43.2 & 51.3 \\
8  & 39.3 & 57.5 & 42.8 & 24.3 & 43.3 & 51.3 \\
9  & 38.9 & 56.5 & 42.0 & 22.9 & 42.4 & 50.3 \\
\hline
\end{tabular}
\label{tab:anchor_scale}
\end{table}

\begin{table}[t]
\centering
\caption{Analysis ($\%$) of different anchor aspect ratios with fixed scale $8S$ on the MS COCO {\tt minival} set.}
\footnotesize \setlength{\tabcolsep}{6.8pt}
\begin{tabular}{|c|ccc|ccc|}
\hline
Aspect Ratio & AP & AP$_{50}$ & AP$_{75}$ & AP$_{\it S}$ & AP$_{\it M}$ & AP$_{\it L}$ \\
\hline
1:4   & 39.1 & 57.2 & 42.3 & 23.1 & 43.1 & 51.4 \\
1:2   & 39.0 & 56.9 & 42.5 & 23.3 & 43.5 & 50.6 \\
1:1   & 39.3 & 57.5 & 42.8 & 24.3 & 43.3 & 51.3 \\
2:1   & 39.3 & 57.4 & 42.3 & 22.8 & 43.4 & 51.0 \\
4:1   & 39.1 & 56.9 & 42.6 & 22.9 & 42.9 & 50.7 \\
\hline
\end{tabular}
\label{tab:anchor_ratio}
\end{table}

\subsection{Comparison}

We compare our final models on the MS COCO {\tt test-dev} subset in Table \ref{tab:coco} with other state-of-the-art object detectors. Following previous works \cite{DBLP:conf/iccv/LinPRK17, DBLP:conf/iccv/abs-1904-01355}, the multi-scale training strategy is adopted for these experiments, \ie, randomly selecting a scale between $640$ to $800$ to resize the shorter side of images during training. Besides, we double the total number of iterations to $180K$ and the learning rate reduction points to $120K$ and $160K$ correspondingly. Other settings are consistent with those mentioned before.

As shown in Table \ref{tab:coco}, our method with ResNet-101 achieves $43.6\%$ AP without any bells and whistles, which is better than all the methods with the same backbone including Cascade R-CNN \cite{DBLP:conf/cvpr/abs-1712-00726} ($42.8\%$ AP), C-Mask RCNN \cite{DBLP:conf/eccv/ChenHT18} ($42.0\%$ AP), RetinaNet \cite{DBLP:conf/iccv/LinPRK17} ($39.1\%$ AP) and RefineDet \cite{DBLP:conf/cvpr/abs-1711-06897} ($36.4\%$ AP). We can further improve the AP accuracy of the proposed method to $45.1\%$ and $45.6\%$ by using larger backbone networks ResNeXt-32x8d-101 and ResNeXt-64x4d-101 \cite{DBLP:journals/corr/XieGDTH16}, respectively. The $45.6\%$ AP result surpasses all the anchor-free and anchor-based detectors except only $0.1\%$ lower than SNIP \cite{DBLP:journals/corr/abs-1711-08189} ($45.7\%$ AP), which introduces the improved multi-scale training and testing strategy. Since our method is about the definition of positive and negative samples, it is compatible and complementary to most of current technologies. We further use the Deformable Convolutional Networks (DCN) \cite{DBLP:conf/iccv/DaiQXLZHW17} to the ResNet and ResNeXt backbones as well as the last layer of detector towers. DCN consistently improves the AP performances to $46.3\%$ for ResNet-101, $47.7\%$ for ResNeXt-32x8d-101 and $47.7\%$ for ResNeXt-64x4d-101, respectively. The best result $47.7\%$ is achieved with single-model and single-scale testing, outperforming all the previous detectors by a large margin. Finally, with the multi-scale testing strategy, our best model achieves $50.7\%$ AP.

\begin{table}[t]
\centering
\caption{Results ($\%$) with different multiple anchors per location on the MS COCO {\tt minival} set.}
\footnotesize \setlength{\tabcolsep}{3.25pt}
\begin{tabular}{|l|cc|ccc|ccc|}
\hline
Method & \#sc & \#ar & AP & AP$_{50}$ & AP$_{75}$ & AP$_{\it S}$ & AP$_{\it M}$ & AP$_{\it L}$ \\
\hline
RetinaNet (\#A=9) & 3 & 3 & 36.3 & 55.2 & 38.8 & 19.8 & 39.8 & 48.8 \\
+Imprs.           & 3 & 3 & 38.4 & 56.2 & 41.6 & 22.2 & 42.4 & 50.1 \\
+Imprs.+ATSS      & 3 & 3 & 39.2 & 57.6 & 42.7 & 23.8 & 42.8 & 50.9 \\
+Imprs.+ATSS      & 3 & 1 & 39.3 & 57.7 & 42.6 & 23.8 & 43.5 & 51.2 \\
+Imprs.+ATSS      & 1 & 3 & 39.2 & 57.1 & 42.5 & 23.2 & 43.1 & 50.3 \\
+Imprs.+ATSS      & 1 & 1 & 39.3 & 57.5 & 42.8 & 24.3 & 43.3 & 51.3 \\
\hline
\end{tabular}
\label{tab:discussion}
\end{table}

\begin{table*}[t]
\renewcommand\arraystretch{0.955}
\centering
\caption{Detection results ($\%$) on MS COCO {\tt test-dev} set. Bold fonts indicate the best performance.}
\vspace{-2mm}
\footnotesize \setlength{\tabcolsep}{9.5pt}
\begin{threeparttable}
\begin{tabular}{|c|c|c|ccc|ccc|}
\hline
Method &Data &Backbone &AP &AP$_{50}$ &AP$_{75}$ &AP$_{\it S}$ &AP$_{\it M}$ &AP$_{\it L}$\\
\hline
\textit{anchor-based two-stage:} & & & & & & & & \\
MLKP \cite{DBLP:journals/corr/abs-1804-00428} &trainval35 &ResNet-101 &28.6 &52.4 &31.6 &10.8 &33.4 &45.1 \\
R-FCN \cite{DBLP:conf/nips/DaiLHS16} &trainval &ResNet-101 &29.9 &51.9 &- &10.8 &32.8 &45.0\\
CoupleNet \cite{DBLP:conf/iccv/abs-1708-02863} &trainval &ResNet-101 &34.4 &54.8 &37.2 &13.4 &38.1 &50.8 \\
TDM \cite{DBLP:journals/corr/ShrivastavaSMG16} &trainval &Inception-ResNet-v2-TDM &36.8 &57.7 &39.2 &16.2 &39.8 &52.1 \\
Hu et al. \cite{DBLP:journals/corr/abs-1711-11575} &trainval35k &ResNet-101 &39.0 &58.6 &42.9 &- &- &- \\
DeepRegionlets \cite{DBLP:conf/eccv/XuLWRBC18} &trainval35k &ResNet-101 &39.3 &59.8 &- &21.7 &43.7 &50.9 \\
FitnessNMS \cite{DBLP:journals/corr/abs-1711-00164} &trainval &DeNet-101 &39.5 &58.0 &42.6 &18.9 &43.5 &54.1 \\
Gu et al. \cite{DBLP:conf/eccv/GuHWWD18} &trainval35k &ResNet-101 &39.9 &63.1 &43.1 &22.2 &43.4 &51.6 \\
DetNet \cite{DBLP:journals/corr/abs-1804-06215} &trainval35k &DetNet-59 &40.3 &62.1 &43.8 &23.6 &42.6 &50.0 \\
Soft-NMS \cite{DBLP:conf/iccv/BodlaSCD17}  &trainval  &ResNet-101     &40.8 &62.4 &44.9 &23.0 &43.4 &53.2 \\
SOD-MTGAN \cite{DBLP:conf/eccv/BaiZDG18} &trainval35k &ResNet-101 &41.4 &63.2 &45.4 &24.7 &44.2 &52.6 \\
G-RMI \cite{DBLP:conf/cvpr/HuangRSZKFFWSG016} &trainval35k &Ensemble of Five Models &41.6 &61.9 &45.4 &23.9 &43.5 &54.9 \\
C-Mask RCNN \cite{DBLP:conf/eccv/ChenHT18} &trainval35k &ResNet-101 &42.0 &62.9 &46.4 &23.4 &44.7 &53.8 \\
Cascade R-CNN \cite{DBLP:conf/cvpr/abs-1712-00726} &trainval35k &ResNet-101 &42.8 &62.1 &46.3 &23.7 &45.5 &55.2 \\
Revisiting RCNN \cite{DBLP:conf/eccv/ChengWSFXH18} &trainval35k &ResNet-101+ResNet-152 &43.1 &66.1 &47.3 &25.8 &45.9 &55.3 \\
SNIP \cite{DBLP:journals/corr/abs-1711-08189} &trainval35k &DPN-98 &45.7 &67.3 &51.1 &29.3 &48.8 &57.1 \\
\hline
\hline
\textit{anchor-based one-stage:} & & & & & & & & \\
YOLOv2 \cite{DBLP:conf/cvpr/RedmonF17} &trainval35k &DarkNet-19 &21.6 &44.0 &19.2 &5.0 &22.4 &35.5\\
SSD512$^*$ \cite{DBLP:conf/eccv/LiuAESRFB16} &trainval35k &VGG-16 &28.8 &48.5 &30.3 &10.9 &31.8 &43.5\\
STDN513 \cite{zhou2018scale}  &trainval &DenseNet-169 &31.8 &51.0 &33.6 &14.4 &36.1 &43.4\\
DES512 \cite{DBLP:journals/corr/abs-1712-00433}  &trainval35k &VGG-16 &32.8 &53.2 &34.5 &13.9 &36.2 &47.5 \\
DSSD513 \cite{DBLP:journals/corr/FuLRTB17} &trainval35k &ResNet-101 &33.2 &53.3 &35.2 &13.0 &35.4 &51.1 \\
RFB512-E \cite{DBLP:conf/eccv/LiuHW18}  &trainval35k  &VGG-16 &34.4 &55.7 &36.4 &17.6 &37.0 &47.6\\
PFPNet-R512 \cite{DBLP:conf/eccv/KimKSKK18}  &trainval35k &VGG-16 &35.2 &57.6 &37.9 &18.7 &38.6 &45.9 \\
RefineDet512 \cite{DBLP:conf/cvpr/abs-1711-06897}  &trainval35k &ResNet-101 &36.4 &57.5 &39.5 &16.6 &39.9 &51.4 \\
RetinaNet \cite{DBLP:conf/iccv/LinPRK17} &trainval35k &ResNet-101 &39.1 &59.1 &42.3 &21.8 &42.7 &50.2 \\
\hline
\hline
\textit{anchor-free keypoint-based:} & & & & & & & & \\
ExtremeNet \cite{DBLP:conf/cvpr/ZhouZK19}   &trainval35k &Hourglass-104 &40.2 &55.5 &43.2 &20.4 &43.2 &53.1 \\
CornerNet \cite{DBLP:conf/eccv/LawD18}   &trainval35k &Hourglass-104 &40.5 &56.5 &43.1 &19.4 &42.7 &53.9 \\
CenterNet-HG \cite{DBLP:journals/corr/abs-1904-07850}   &trainval35k &Hourglass-104 &42.1 &61.1 &45.9 &24.1 &45.5 &52.8 \\
Grid R-CNN \cite{DBLP:conf/cvpr/LuLYLY19}   &trainval35k &ResNeXt-101 &43.2 &63.0 &46.6 &25.1 &46.5 &55.2 \\
CornerNet-Lite \cite{DBLP:journals/corr/abs-1904-08900}   &trainval35k &Hourglass-54 &43.2 &- &- &24.4 &44.6 &57.3 \\
CenterNet \cite{DBLP:journals/corr/abs-1904-08189}   &trainval35k &Hourglass-104 &44.9 &62.4 &48.1 &25.6 &47.4 &57.4 \\
RepPoints \cite{DBLP:journals/corr/abs-1904-11490}   &trainval35k &ResNet-101-DCN &45.0 &66.1 &49.0 &26.6 &48.6 &57.5 \\
\hline
\hline
\textit{anchor-free center-based:} & & & & & & & & \\
GA-RPN \cite{DBLP:conf/cvpr/WangCYLL19} &trainval35k &ResNet-50 &39.8 &59.2 &43.5 &21.8 &42.6 &50.7 \\
FoveaBox \cite{DBLP:journals/corr/abs-1904-03797} &trainval35k &ResNeXt-101 &42.1 &61.9 &45.2 &24.9 &46.8 &55.6 \\
FSAF \cite{DBLP:conf/cvpr/ZhuHS19} &trainval35k &ResNeXt-64x4d-101 &42.9 &63.8 &46.3 &26.6 &46.2 &52.7 \\
FCOS \cite{DBLP:conf/iccv/abs-1904-01355} &trainval35k &ResNeXt-64x4d-101 &43.2 &62.8 &46.6 &26.5 &46.2 &53.3 \\
\hline
\hline
\textit{Ours:} & & & & & & & & \\
ATSS &trainval35k &ResNet-101 &43.6 &62.1 &47.4 &26.1 &47.0 &53.6 \\
ATSS &trainval35k &ResNeXt-32x8d-101 &45.1 &63.9 &49.1 &27.9 &48.2 &54.6 \\
ATSS &trainval35k &ResNeXt-64x4d-101 &45.6 &64.6 &49.7 &28.5 &48.9 &55.6 \\
ATSS &trainval35k &ResNet-101-DCN &46.3 &64.7 &50.4 &27.7 &49.8 &58.4 \\
ATSS &trainval35k &ResNeXt-32x8d-101-DCN &47.7 &66.6 &52.1 &29.3 &50.8 &59.7 \\
ATSS &trainval35k &ResNeXt-64x4d-101-DCN &47.7 &66.5 &51.9 &29.7 &50.8 &59.4 \\
ATSS (Multi-scale testing) &trainval35k &ResNeXt-32x8d-101-DCN &50.6 &68.6 &56.1 &\bf{33.6} &\bf{52.9} &62.2 \\
ATSS (Multi-scale testing) &trainval35k &ResNeXt-64x4d-101-DCN &\bf{50.7} &\bf{68.9} &\bf{56.3} &33.2 &\bf{52.9} &\bf{62.4} \\
\hline
\end{tabular}
\end{threeparttable}
\label{tab:coco}
\end{table*}

\subsection{Discussion}
Previous experiments are based on RetinaNet with only one anchor per location. There is still a difference between anchor-based and anchor-free detectors that is not explored: the number of anchors tiled per location. Actually, the original RetinaNet tiles $9$ anchors ($3$ scales $\times$ $3$ aspect ratios) per location (marked as \textbf{RetinaNet (\#A=9)}) that achieves $36.3\%$ AP as listed in the first row of Table \ref{tab:discussion}. In addition, those universal improvements in Table \ref{tab:inconsistency} can also be used to RetinaNet (\#A=9), boosting the AP performance from $36.3\%$ to $38.4\%$. Without using the proposed ATSS, the improved RetinaNet (\#A=9) has better performance than RetinaNet (\#A=1), \ie, $38.4\%$ in Table \ref{tab:discussion} \emph{vs.} $37.0\%$ in Table \ref{tab:inconsistency}. These results indicate that under the traditional IoU-based sample selection strategy, tiling more anchor boxer per location is effective.

However, after using our proposed method, the opposite conclusion will be drawn. To be specific, the proposed ATSS also improves RetinaNet (\#A=9) by $0.8\%$ on AP, $1.4\%$ on $AP_{50}$ and $1.1\%$ on $AP_{75}$, achieving similar performances to RetinaNet (\#A=1) as listed in the third and sixth rows of Table \ref{tab:discussion}. Besides, when we change the number of anchor scales or aspect ratios from $3$ to $1$, the results are almost unchanged as listed in the fourth and fifth rows of Table \ref{tab:discussion}. In other words, as long as the positive samples are selected appropriately, no matter how many anchors are tiled at each location, the results are the same. We argue that tiling multiple anchors per location is a useless operation under our proposed method and it needs further study to discover its right role.

\section{Conclusion}

In this work, we point out that the essential difference between one-stage anchor-based and center-based anchor-free detectors is actually the definition of positive and negative training samples. It indicates that how to select positive and negative samples during object detection training is critical. Inspired by that, we delve into this basic issue and propose the adaptive training sample selection, which automatically divides positive and negative training samples according to statistical characteristics of object, hence bridging the gap between anchor-based and anchor-free detectors. We also discuss the necessity of tiling multiple anchors per location and show that it may not be a so useful operation under current situations. Extensive experiments on the challenging benchmarks MS COCO illustrate that the proposed method can achieve state-of-the-art performances without introducing any additional overhead.

\section*{Acknowledgments}
This work has been partially supported by the Chinese National Natural Science Foundation Projects $\#61872367$, $\#61876178$, $\#61806196$, $\#61806203$, $\#61976229$.

\newpage
{\small
\bibliographystyle{ieee_fullname}
\bibliography{reference}
}

\end{document}